\documentclass[UTF8]{article}

\PassOptionsToPackage{numbers, compress}{natbib}

\usepackage[preprint]{neurips_2023}

\usepackage[utf8]{inputenc} 
\usepackage[T1]{fontenc}    
\usepackage[pagebackref,breaklinks,colorlinks]{hyperref}
\usepackage{url}            
\usepackage{booktabs}       
\usepackage{amsfonts}       
\usepackage{nicefrac}       
\usepackage{microtype}      
\usepackage{xcolor}         
\usepackage{CJKutf8}

\usepackage{amsmath}
\usepackage{amsfonts}
\usepackage{amssymb}
\usepackage{wrapfig}
\usepackage{subcaption}
\usepackage{multirow}
 \usepackage{mathtools} 

\usepackage{verbatim}

\usepackage{anyfontsize}

\usepackage{microtype}
\usepackage{graphicx}
\usepackage{booktabs} 
\usepackage{multirow}
\usepackage{amsmath,amssymb}
\usepackage{booktabs}
\usepackage{caption,subcaption}

\usepackage{xcolor}
\definecolor{mygreen}{HTML}{3cb44b}
\definecolor{skyblue}{HTML}{beffff}
\definecolor{lightgreen}{HTML}{90ee90}

\usepackage{color, colortbl}

\definecolor{emerald}{rgb}{0.31, 0.78, 0.37}

\usepackage{tcolorbox}
\usepackage{enumitem}
\setitemize{itemsep=10pt,topsep=0pt,parsep=0pt,partopsep=0pt}
\usepackage{colortbl}

\usepackage{xcolor}
\definecolor{mygreen}{HTML}{3cb44b}
\colorlet{myyellow}{green!10!orange!90!}
\makeatletter

\usepackage{tikz}
\usetikzlibrary{arrows,shapes,snakes,automata,backgrounds,fit,petri}
\usepackage{adjustbox}

\newcommand{\RN}[1]{%
	\textup{\lowercase\expandafter{\it \romannumeral#1}}%
}
\usepackage{tabu}

\newcommand{\beq}{\vspace{0mm}\begin{equation}}
\newcommand{\eeq}{\vspace{0mm}\end{equation}}
\newcommand{\beqs}{\vspace{0mm}\begin{eqnarray}}
\newcommand{\eeqs}{\vspace{0mm}\end{eqnarray}}
\newcommand{\barr}{\begin{array}}
\newcommand{\earr}{\end{array}}

\usepackage{color, colortbl}
\definecolor{Gray}{gray}{0.93}

\usepackage{lipsum}

\usepackage{pifont}

\usepackage{makecell}

\usepackage{xcolor,amsmath}
\usepackage[linesnumbered,ruled,vlined]{algorithm2e}
\DontPrintSemicolon

\usepackage{xcolor}
\definecolor{mygreen}{HTML}{3cb44b}

\SetKwComment{Comment}{\color{green!50!black}\# }{}

\SetKwProg{Function}{def}{:}{}

\SetKwProg{For}{for}{:}{}
\SetKwProg{If}{if}{:}{}

\usepackage{xcolor}

\newcommand{\shortname}{X-LLM}

\title{X-LLM: \\Bootstrapping Advanced Large Language Models by Treating Multi-Modalities as Foreign Languages}

\author{
Feilong Chen$^{1,2}$, Minglun Han$^{1,3}$, \\ \bf{Haozhi Zhao}$^{1,3}$, Qingyang Zhang$^{1,2}$, Jing Shi$^{1}$, Shuang Xu$^{1}$ and Bo Xu$^{1,2,3}$ \\
$^{1}$Institute of Automation, Chinese
Academy of Sciences  \\
$^2$School of Future Technology, University of Chinese
Academy of Sciences \\
$^3$School of Artificial Intelligence, University of Chinese Academy of Sciences \\
\href{https://x-llm.github.io}{https://x-llm.github.io}
}

\begin{document}

\maketitle

\begin{abstract}
Large language models (LLMs) have demonstrated remarkable language abilities. GPT-4, based on advanced LLMs, exhibits extraordinary multimodal capabilities beyond previous visual language models. We attribute this to the use of more advanced LLMs compared with previous multimodal models. Unfortunately, the model architecture and training strategies of GPT-4 are unknown. To endow LLMs with multimodal capabilities, we propose X-LLM, which converts Multi-modalities (images, speech, videos) into foreign languages using X2L interfaces and inputs them into a large Language model (ChatGLM). Specifically, X-LLM aligns multiple frozen single-modal encoders and a frozen LLM using X2L interfaces, where ``X'' denotes multi-modalities such as image, speech, and videos, and ``L'' denotes languages. X-LLM's training consists of three stages: (1) Converting Multimodal Information: The first stage trains each X2L interface to align with its respective single-modal encoder separately to convert multimodal information into languages. (2)  Aligning X2L representations with the LLM: single-modal encoders are aligned with the LLM through X2L interfaces independently. (3) Integrating multiple modalities: all single-modal encoders are aligned with the LLM  through X2L interfaces to integrate multimodal capabilities into the LLM. 
Our experiments show that X-LLM demonstrates impressive multimodel chat abilities, sometimes exhibiting the behaviors of multimodal GPT-4 on unseen images/instructions, and yields a 84.5\% relative score compared with GPT-4 on a synthetic multimodal instruction-following dataset. And we also conduct quantitative tests on using LLM for ASR and multimodal ASR, hoping to promote the era of LLM-based speech recognition.

\end{abstract}

\section{Introduction}
In recent years, multimodal language models~\cite{li2019visualbert,li2023blip2,huang2023language} have undergone rapid development. These models possess excellent abilities in multimodal understanding and response generation and can perform well in tasks such as image captioning~\cite{vinyals2016show}, visual question answering~\cite{antol2015vqa}, visual dialog~\cite{das2017visual}, video captioning~\cite{gao2017video}, and spoken dialogue~\cite{wen2015semantically}. It is worth noting that a large-scale multimodal model, GPT-4~\cite{openaigpt4}, has recently been introduced, demonstrating many impressive capabilities. For example, GPT-4 can follow various instructions to complete language tasks, and can also answer various questions about images. For instance, GPT-4 can give detailed and accurate descriptions of images, understand and explain the humor in visual content, and even provide correct website-building code based on handwritten code images. Although GPT-4 demonstrates remarkable capabilities, unfortunately, we do not know the details of its model structure and training methods. We believe that this is due to the fact that GPT-4 uses a more advanced and larger language model compared to previous multimodal models. With the support of powerful language abilities, GPT-4 can express understood visual content in the form of language.

To validate this hypothesis and endow LLM with multimodal capabilities, we propose X-LLM. It converts multimodal information, such as images, speech, and videos, into foreign languages using X2L interfaces, and then feeds converted multimodal information into a large language model (ChatGLM). Specifically, X-LLM aligns multiple frozen single-modal encoders and a frozen LLM using X2L interfaces. X2L interfaces consist of an image I2L interface, a video V2L interface, and a speech S2L interface, where ``X'' denotes the multi-modalities and ``L'' denotes languages. The image interface and video interface have the same structure, and we adopt the Q-Former from BLIP-2~\cite{li2023blip2} to convert visual information into foreign language representations. For efficiency, the video interface reuses the parameters of the image interface with image-text data but is further trained with video-text data to align the encoded video features with the LLM. The speech interface utilizes the continuous integrate-and-fire (CIF) mechanism~\cite{Dong020,HanDZX21} and transformer structure to convert speech utterance into foreign language representations. The training of X-LLM consists of three stages. (1) Converting Multimodal Information: the first stage trains each X2L interface to align with its respective single-modal encoder separately to convert multimodal information into languages. (2) Aligning X2L representations with the LLM: single-modal encoders are aligned with the LLM through X2L interfaces. (3) Integrating multiple modalities: all single-modal encoders are aligned with the LLM  through X2L interfaces to integrate multimodal capabilities into the LLM. In the first two stages, we use image caption data, video caption data and automatic speech recognition (ASR) data to train the X2L interfaces. To better equip LLM with multimodal capabilities, we construct a multimodal instruction dataset ($\sim$10K) based on open-source datasets to further improve the proposed model. Although without the third training stage, X-LLM already has the ability to accomplish multimodal tasks such as visual spoken question answering, we find that with only rare additional multimodal instruction data, LLM can further unify the capabilities of multiple modalities.

In our experiments, we find that X-LLM has abilities similar to those of GPT-4. For example, X-LLM can generate complex image descriptions and explain unusual visual phenomena. In our research, when using input images, X-LLM can recognize the location in the image, such as identifying the Forbidden City and providing relevant information about it, observe the food in the image and provide detailed recipes; create stories for pictures, and come up with textual meanings for logos. We also find that X-LLM's image-related abilities can be extended to videos, such as introducing the content of a video, retrieving movie names, or art-related facts directly from the video. Moreover, X-LLM can answer questions based on spoken questions and can combine images or videos to answer spoken questions. These abilities are previously not present in previous multimodal models but are now made possible by the powerful language modeling capabilities of X-LLM.

Our contributions are summarised as follows:
\begin{itemize}[leftmargin=7.5mm]
\setlength{\itemsep}{2pt}
\item 
{\it Multimodal LLM framework}. We propose X-LLM, a Multimodal LLM which injects multiple modalities (such as images, speech, and videos) into LLM through X2L interfaces, giving LLM the ability to process multimodal data. This framework has good scalability and can be extended to more modalities. ``X'' in X2L interfaces can be any  modality. We compare our X-LLM with LLaVA and MiniGPT-4 in terms of the ability to handle visual inputs with Chinese elements, and find that X-LLM outperformed them significantly. We also conduct quantitative tests on using LLM for ASR and multimodal ASR, hoping to promote the era of LLM-based speech recognition.

\item
{\it Transferability of parameters in English image-text alignment modules}. We find that the Q-former module trained on English image-text data can be transferred to other languages. In our experiments, we have successfully transferred the model parameters from Indo-European English to Sino-Tibetan Chinese. The transferability of language greatly increases the possibility of using English image-text data and its trained model parameters, and improves the efficiency of training multimodal LLMs in other languages.

\item
{\it Open-source}. We construct a concise and high-quality Chinese multimodal instruction dataset. By training X-LLM on this multimodal instruction data, X-LLM can better integrate the multimodal capabilities acquired through multiple encoders and corresponding X2L interfaces. And  We release the following assets to the public: the generated multimodal instruction data, the codebase for model training, the model checkpoint, and a multimodal chat demo. 

\end{itemize}

\section{Related Work}

\begin{figure}[t]
\centering  
\includegraphics[width=0.99\textwidth]{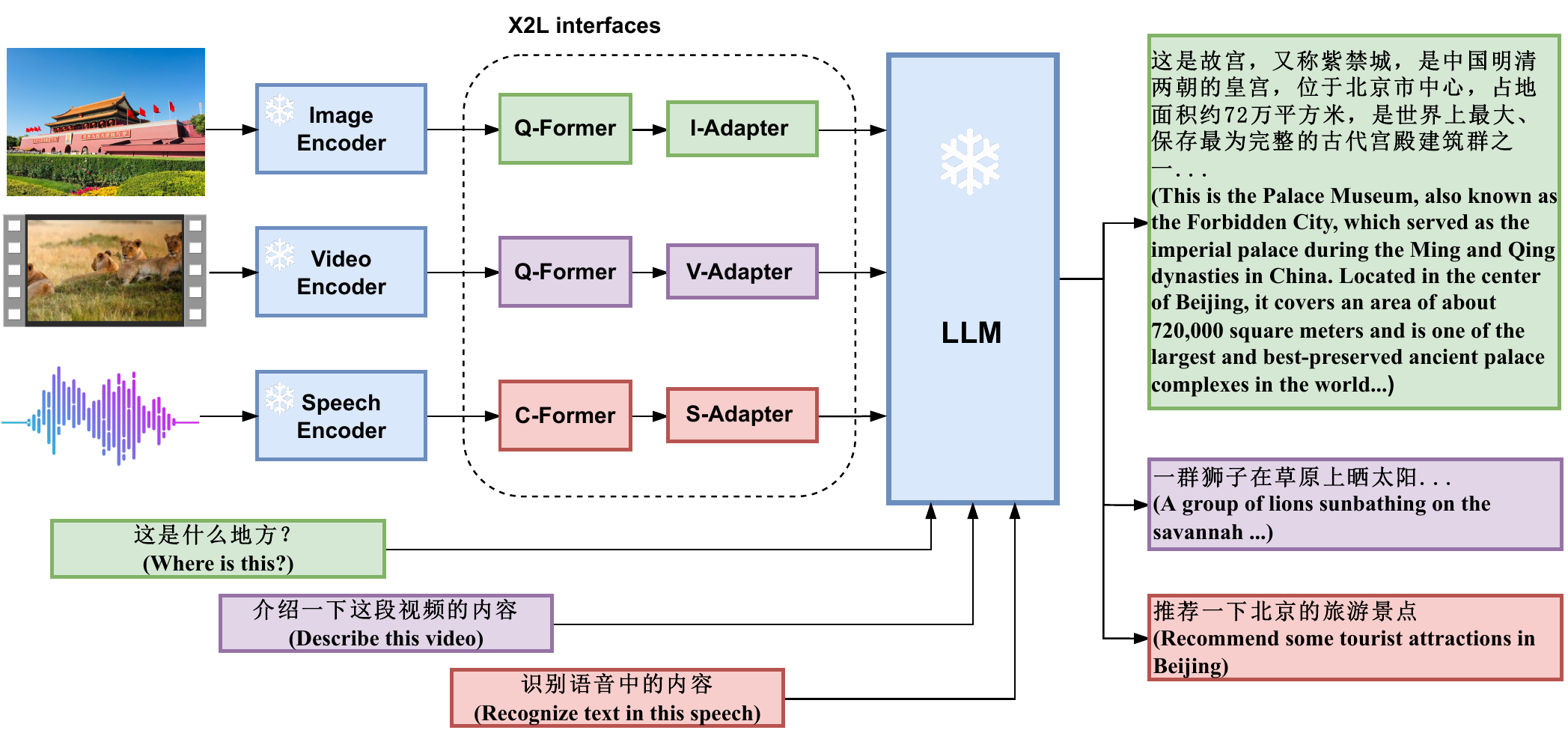}
\caption{\shortname{} network architecture.}
\label{fig:xllm_arch}  \vspace{-10pt}
\end{figure}

\paragraph{Vision-Language Models.}
As summarized in many surveys~\cite{chen2023vlp,du2022survey}, visual language models~\cite{zhou2020unified,li2022blip} have made great strides with the development of pre-training techniques~\cite{devlin2018bert,bi2020palm,zoph2020rethinking}. In the early days, researchers used Faster-RCNN~\cite{girshick2015fast} to extract image features and concatenated them with language models such as BERT~\cite{devlin2018bert} to perform vision-language pre-training. VisualBERT~\cite{li2019visualbert}, for example, combines image regions and language using a Transformer~\cite{vaswani2017attention} to allow self-attention to discover implicit alignments between language and vision. It is pre-trained with masked language modeling~\cite{devlin2018bert} and a sentence-image prediction task~\cite{li2019visualbert}. With the introduction of ViLT~\cite{kim2021vilt}, researchers use vision transformers~\cite{dosovitskiy2020image,khan2022transformers} to process images, textual transformers (such as BERT~\cite{devlin2018bert}, GPT-2~\cite{radford2019language}, T5~\cite{raffel2020exploring}) to process text, and pre-training objectives such as masked language modeling, image-text matching, and image-text contrast to train visual language models. CLIP~\cite{radford2021learning} uses a text encoder and an image encoder to encode text and images separately and then performs unsupervised contrastive learning to obtain good representations of vision-language alignment. BLIP~\cite{li2022blip} is a new VLP framework that transfers flexibly to both vision-language understanding and generation tasks.


In the field of visual dialogue~\cite{das2017visual,chen2020dmrm,chen2022utc}, researchers design pre-training objectives related to visual dialogue based on vision-language pre-training models~\cite{devlin2018bert,li2019visualbert} and finetune vison-language models on visual dialogue data~\cite{das2017visual} to achieve better dialogue performance. VisDial-BERT~\cite{murahari2020large} and VD-BERT~\cite{wang2020vd}, for example, use pre-trained ViLBERT~\cite{lu2019vilbert} and BERT to finetune models on visual dialogue data using masked language modeling and image-text matching. AlignVD~\cite{chen2022unsupervised} proposes two methods for visual-language alignment based on pre-trained ViT~\cite{radford2021learning} and BERT to achieve better performance in visual dialogue. \vspace{-10pt}

\paragraph{Enhancing Vision-language Understanding with Advanced LLMs.}
Although the aforementioned vision-language models have achieved some success, there is still significant room for improvement in terms of language generation~\cite{chen2023vlp,long2022vision,li2023blip2}. A recent method~\cite{li2023blip2,driess2023palm,openaigpt4} for enhancing visual language understanding using advanced large-scale language models~\cite{touvron2023llama,chiang2023vicuna} has been proposed. For example, BLIP2~\cite{li2023blip2} uses a Q-Former to connect a visual encoder with an LLM, aligning the learned queries of the Q-Former with language-related visual features extracted by the visual encoder. The Q-Former then connects the visual encoder with the language model, allowing the learned query representations to adapt to the LLM. PaLM-E~\cite{driess2023palm} combines ViT-22B~\cite{dehghani2023scaling} with PaLM-560B~\cite{bi2020palm} to inject multimodal information into the embedding space of the pre-trained language model, establishing a connection between perception and language and greatly enhancing the model's visual language understanding ability. In addition, Visual ChatGPT~\cite{wu2023visual} and HuggingGPT~\cite{shen2023hugginggpt} use ChatGPT as the core logic controller, which understands user intent and then call upon specific domain visual language models. Finally, the recently proposed GPT-4~\cite{openaigpt4} demonstrates powerful multimodal capabilities: building on its strong language understanding abilities, it can generate complex image descriptions, create websites based on handwritten text instructions, and explain unusual visual phenomena. However, the model structure and training strategies of GPT-4 remain a mystery. MiniGPT-4~\cite{zhu2023minigpt4} and LLaVA~\cite{liu2023visual} align text and image data to the large-scale language model Vicuna~\cite{chiang2023vicuna} and ViT~\cite{zhai2022scaling} to complete image-based language tasks. In contrast, X-LLM is a universal framework for multimodal LLMs that bootstraps advanced large language models by treating multi-modalities as foreign languages. In this paper, we implement X-LLM that supports images, videos, and speech. Based on the X-LLM framework, we can extend the model to more modalities, such as injecting continuous space robot states, terminal information, or audio rather than speech into the LLM.
\section{Approach}
X-LLM aims to align multiple pre-trained single-modal encoders with advanced large-scale language models (LLMs), as shown in Figure~\ref{fig:xllm_arch}. Specifically, we use ChatGLM~\footnote{\url{https://github.com/THUDM/ChatGLM-6B}} as the language decoder, which is built on top of GLM~\cite{du2022glm,zeng2023glm-130b} and can perform various complex language tasks. For visual perception, we adopt ViT-g~\cite{zhai2022scaling}, as the image encoder and video encoder. For speech perception, we use a speech encoder comprised of convolution layers and conformer structure~\cite{gulati2020conformer}. We design a module that aligns multimodal information with LLM, collectively referred to as the X2L interfaces, which includes an image interface, a video interface, and a speech interface. The image interface and the video interface have the same structure which consists of Q-Formers~\cite{li2023blip2} and Adapter modules. The speech interface includes the C-Former and an Adapter module. The C-Former could compress the frame-level speech feature sequence from the speech encoder into the token-level speech embedding sequence with continuous integrate-and-fire (CIF) mechanism~\cite{Dong020,HanDZX21,HanDLCZMX22}. As the token-level speech embedding sequence is strictly aligned with the token sequence of the transcription corresponding to the speech utterance, representing speech using token-level speech embeddings can effectively reduce the GPU memory usage when incorporating speech into LLMs. 

\subsection{X2L Interfaces}
X2L interfaces aim to convert multimodal information into foreign languages, which includes an image interface, a video interface, and a speech interface.

\paragraph{The Image Interface.}
Inspired by~\cite{li2023blip2}, the image interface consists of a Q-Formers~\cite{li2023blip2} and an I-Adapter module. The Q-Formers aims to convert images into languages, where image features obtained from the image encoder are converted into a sequence with $L_i$ quasi-linguistic embeddings. The I-Adapter module aims to align the dimensions of the quasi-linguistic embeddings and the embedding dimension of the LLM. 

\paragraph{The Video Interface.}
The video interface has the same structure as the image interface, which also consists of Q-Formers~\cite{li2023blip2} and a V-Adapter module. We use uniform sampling and represent each video with $T$ frames. We then treat each frame as an image. The video interface converts each frame features into a sequence with $L_i$ quasi-linguistic embeddings. Then the video interface concatenates all the sequences to obtain the final quasi-linguistic embeddings, which have a length of $T \times L_i$.  

\paragraph{The Speech Interface.}
To transform the speech features from the speech encoder into more semantic representations, we introduce a speech-to-language interface called the speech interface. The speech interface consists of two parts, namely the C-Former and the S-Adaptor. The C-Former is the combination of a CIF module and a 12-layer transformer structure~\cite{devlin2018bert}. First, the CIF module compresses the speech feature sequence from the speech encoder into a token-level speech embedding sequence with the same length as the corresponding transcription via variable-length down-sampling. Assuming the length of the feature sequence emitted by the speech encoder for the input speech is $U$, and the length of the token sequence of the transcription of the speech utterance is $L_{s}$, the length of the token-level speech embedding sequence should be $L_{s}$ ($U$ is usually several times longer than $L_{s}$). Then, the transformer structure provides contextual modeling for the token-level speech embeddings from the CIF module. Finally, the S-Adaptor is used to project the outputs of the transformer structure to the input vector space of the LLM, further narrowing down the semantic gap between speech and language.

\subsection{Training Strategy}
To efficiently implement X-LLM, we propose a three-stage training strategy. (1) Converting Multimodal Information: we align the Image Encoder with the Q-Former of the image (green part), and the Speech Encoder with the CIF module. (2) Aligning X2L representations with the LLM: in the second stage, we align the Image Encoder with the LLM through the image interface, align the Video Encoder with the LLM through the video interface, and align the Speech Encoder with LLM through the speech interface. In the third stage, we integrate training of the image, video, and speech, and align the overall single-modal encoders with the LLM using a smaller but high-quality multimodal instruction dataset (such as instructions containing visual spoken dialogue, i.e., responding to spoken dialogue inputs based on images).

\subsubsection{First Training Stage: Converting Multimodal Information}\label{sec:firststage}
In the first stage, the traditional approach is to align the Image Encoder with the image Q-Former using a large amount of image-text data, similar to the first stage of BLIP2~\cite{li2023blip2} which utilized around 500 million image-text pairs. However, we find that while BLIP2 used English data, we can still leverage the pretrained parameters of the Q-Former in BLIP2 to implement a Chinese Multimodal LLM. Therefore, in the first stage, to efficiently implement X-LLM, we only convert the representation of the speech encoder to a quasi-linguistic representation through the speech interface.

For the speech-related structures, we train a CIF-based ASR model with multiple ASR datasets containing to obtain the speech encoder and CIF module in the C-Former. The CIF-based ASR model consists of a speech encoder, a CIF module, and a decoder~\cite{Dong020}. We employ the speech encoder of this ASR model as the speech encoder and employ the CIF module of this ASR model as that in the C-Former of the speech interface. Note that the parameters of the speech encoder and CIF module are kept frozen during all subsequent training stages. Please refer to the appendix for more details about the structure and training of the CIF-based ASR model.

\subsubsection{Second Training Stage: Aligning X2L Representations with the LLM}
As mentioned above, despite the difference in language, we are still able to reuse the parameters of the Q-Former in BLIP2. Specifically, we used the Q-Former trained in the second stage of BLIP2 to initialize the image interface's Q-Former in X-LLM. To adapt the Q-Former to Chinese LLM, we use a combined dataset, totaling approximately 14 million Chinese image-text pairs for training. 


Next, we use the trained image interface to initialize the video interface (the Q-Former and the V-Adapter) and train the video interface on the translated video-text data.

Finally, we train the speech interface using ASR data to align the output of the speech interface with the LLM. It should be noted that throughout the entire second training stage, all the encoders and the LLM remain frozen, with only the interfaces being trained. 


\subsubsection{Third Training stage: Integrating Multiple Modalities}
After the first two stages of training, our X-LLM has demonstrated a remarkable ability to provide reasonable answers to human queries based on multimodal information and has gained a vast amount of knowledge. We have observed that, even without the instruction for joint training on multiple modalities, such as "answer questions based on images using voice input," X-LLM is capable of performing tasks that require multiple modalities, such as visual spoken dialogue, multimodal speech recognition, and multimodal machine translation. This remarkable ability is likely due to X-LLM's integration of LLM's excellent instruction generalization capability, which has been extended to the multimodal domain. This ability enables us to train more modalities independently in the first two stages and integrate them into the model without the need for joint training with existing modalities.

To explore the potential of multimodal joint instruction data in further enhancing X-LLM's ability to perform multimodal tasks, such as visual spoken question answering, we have constructed a concise but high-quality multimodal instruction dataset. Different from MiniGPT-4~\cite{zhu2023minigpt4} and LLaVA~\cite{liu2023visual}'s datasets, which only contain image-text instruction data and other textual instruction datasets for instruction finetuning and conversations, our dataset supports multimodal joint instructions and includes (1) image-text instruction data, (2) speech-text instruction data, (3) video-text instruction data, and (4) image-text-speech instruction data. 

\paragraph{Constructing a High-quality Alignment Dataset for Multimodal LLM.}\label{instructiondata}
We use ChatGPT to translate 3.5K image-text instructions built by MiniGPT-4. Then, we manually select 2k data from AISHELL-2~\cite{du2018aishell} and write 5 different instructions for speech recognition tasks.  We use ChatGPT to translate the ActivityNet dataset~\cite{krishna2017dense}, followed by manually selecting 1k data and writing 5 different instructions for corresponding video-text tasks. We manually select and rewrite 1k data from self-constructed VSDial-CN data, aiming to enable the model to perform dialogue generation tasks based on images and speech. More details of the data can be found in the appendix, including the details of the training data for the first two stages and the multimodal instruction data.

\paragraph{The Third Training Stage.}
During this stage, we use the constructed compact yet high-quality data to finetune our model. During finetuning, we use the predefined prompts in the following template: 

\textit{<Image><ImageFeats></Image><Video><VideoFeats></Video><Speech><SpeechFeats>
</Speech>Question: <Instruction>$\backslash$n Answer:} 

In this prompt, <Instruction> represents a randomly sampled instruction from our predefined instruction set, including different forms such as ``describe this image in detail'', ``can you describe what you notice in the video'', or ``answer the question in the speech based on the image''. It should be noted that we do not calculate regression loss specifically for this particular instruction prompt. Therefore, X-LLM can integrate multiple modalities and generate more natural and reliable responses based on various combinations of instructions as needed.

\section{Experiments}
\subsection{Multimodal Chat}
We have developed a Chatbot demo to show multimodal understanding and conversation abilities of X-LLM. For comparisons, query LLaVA~\cite{liu2023visual}\footnote{\url{https://llava-vl.github.io/}} and MiniGPT-4~\cite{zhu2023minigpt4}\footnote{\url{https://minigpt-4.github.io/}} from their online demos to get their response.


As shown in Table~\ref{tab:visual_example_forbidden} and~\ref{tab:visual_example_kings}, although LLaVA and MiniGPT-4 also exhibit the characteristic of generating answers based on the given prompt, their answers regarding visual content with Chinese elements are not as satisfactory. In the first example about the Forbidden City shown in Table~\ref{tab:visual_example_forbidden}, X-LLM recognizes that the place is the Forbidden City and provides a detailed introduction to its history, architecture, and style. LLaVA describes Chinese palaces and flags, but it does not recognize that the famous palace is the Forbidden City and therefore cannot provide relevant information about it. MiniGPT-4 exhibits the same problem and tends to describe the image more. In the second example about the game ``Honor of Kings'' shown in Table~\ref{tab:visual_example_kings}, X-LLM identifies it as a multiplayer online battle arena game, ``Honor of Kings'', developed by Tencent and provides accurate release time. LLaVA, on the other hand, gives multiple incorrect answers, as there are no elements of popular games such as snakes and pocket monsters in the image, and the game is not played with a mouse. MiniGPT-4 fails to recognize the game and provides a more generic description.

For video input and speech input, we provide some examples as shown in Appendix~\ref{lab:examples}. 

\paragraph{Quantitative Evaluation.}
In order to systematically evaluate the performance of the X-LLM model on visual input, we aim to use quantitative metrics to measure the model's ability to follow instructions. We adopt an evaluation method similar to that proposed by LLaVA~\cite{liu2023visual} and use ChatGPT to measure the quality of the answers generated by our model. Specifically, we use the LLaVA-test dataset~\cite{liu2023visual} provided by LLaVA, which contains 30 randomly selected images from the COCO validation set, each with three types of questions (conversation, detailed description, and complex reasoning). We first translate the questions into Chinese, and X-LLM predicts the answers based on the translated Chinese questions and visual input images. Then we translate the responses given by X-LLM into English for comparison with GPT-4. GPT-4 makes reference predictions based on the question, ground truth bounding boxes, and captions, marking the upper limit. After obtaining the responses from the two models, we provide the question, visual information (in the form of captions and bounding boxes), and generated responses from both assistants to ChatGPT. ChatGPT evaluates the usefulness, relevance, accuracy, and level of detail of the assistants' responses and gives an overall score from 1 to 10, with higher scores indicating better overall performance. ChatGPT is also required to provide a comprehensive evaluation explanation for a better understanding of the model. LLaVA used GPT-4 as a teacher to evaluate the quality of the responses generated by LLaVA and GPT-4, while we believe that using a non-GPT-4 evaluation model (i.e. using ChatGPT) will be more objective (Also because we do not have GPT-4 API.). Examples of test questions can be found in Appendix~\ref{lab:ques}.

We show the results in Table~\ref{tab:results}. Although different evaluation models are used (LLaVA uses GPT-4, X-LLM uses ChatGPT), we are able to make rough comparisons. The results show that X-LLM yields a performance of 84.5\% nearly GPT-4. X-LLM outperforms LLaVA in terms of conversation and detail description but is inferior in complex reasoning. There are two reasons for this. One reason is that X-LLM do not use the 150k visual instruction dataset proposed by LLaVA, which has the same format as the test set. The second reason is that X-LLM has fewer language model parameters. It is based on ChatGLM with 6B parameters, while LLaVA is based on Vicuna with 13B parameters. And we do not finetune the LLM while LLaVA finetune the LLM Vicuna. 

Furthermore, comparing ``X-LLM w/ 4M'' and ``X-LLM w/ 4M no init'', we can observe that using the BLIP2 pre-trained Q-Former parameters significantly improves the model's performance, which This verifies the transferability of parameters in the English image text alignment module. The transferability of language greatly increases the possibility of using English image-text data and its trained model parameters, and improves the efficiency of training multimodal LLMs in other languages. Comparing X-LLM and ``X-LLM w/ 4M'', we can see that increasing the number of image-text pairs used during training can enhance the model's performance. However, we also notice that X-LLM performs worse than ``X-LLM w/ 4M'' in complex reasoning, which may be attributed to the additional use of the Wukong dataset~\cite{gu2022wukong}, whose quality is inferior to that of the dataset comprising the 4M data. Note that all three variants of X-LLM achieve high performances on complex reasoning which is because of the powerful ability of language models and complex reasoning questions can also be answered without images to some extent.

\begin{table}[t]
\centering
\scalebox{0.87}{
\begin{tabular}{l|cccc}
\toprule
Model & Conversation & Detail description & Complex reasoning & \textbf{All} \\
\midrule
LLaVA  & 83.1 & 75.3 & \bf{96.5} & \bf{85.1} \\
\midrule
X-LLM & \bf{85.4} & \bf{83.5} & 84.6 & 84.5 \\
\quad \quad w/ 4M & 74.8 & 83.7 & 86.5 & 81.9 \\
\quad \quad w/ 4M no init & 64.7 & 71.9 & 85.0 & 73.8 \\
\bottomrule
\end{tabular}
}
\vspace{2mm}
\caption{Relative scores for different settings {\it w.r.t.} GPT-4 (text-only) on 30 randomly sampled images from COCO Val 2014. Each image is associated one short question, one detailed question, and one complex reasoning question, resulting in a total of 90 questions. We prompt ChatGPT with the answers from our model outputs and the answers by GPT-4 (text-only), and let it compare between both responses and give a rating with an explanation. ``w/ 4M'' denotes that we train the image interface only using 4M image-text pairs. ``w/ 4M no init'' denotes that we train the image interface only using 4M image-text pairs and without using the parameters of pretrained BLIP2.}
\label{tab:results}\vspace{-20pt}
\end{table}

\begin{table}
  \begin{minipage}{0.99\textwidth}
\centering  
\vspace{-4mm}
 
\vspace{1mm}
\scalebox{0.88}{
\begin{tabular}{l p{12.5cm} }
\toprule
 \multicolumn{2}{l}{\bf Visual input example, The Forbidden City:}  \\
\midrule
&  \includegraphics[height=3.5cm]{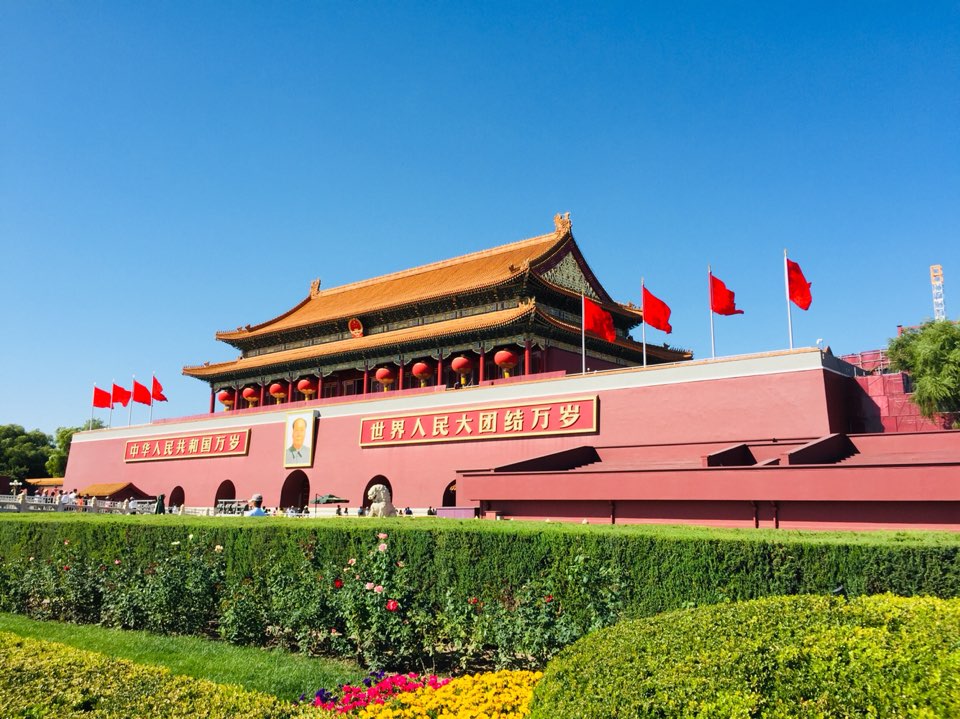} \\
\midrule

User &  \begin{CJK}{UTF8}{gbsn}详细介绍一下这个地方\end{CJK} (Introduce the place in details)\\
\shortname{} & \begin{CJK}{UTF8}{gbsn}
北京的故宫是中国封建王朝皇家宫殿的代表，也是世界上保存最完整的古代宫殿建筑群之一，位于北京市中心紫禁城内，是中国最重要的文化遗产之一。

\quad 

故宫始建于明朝永乐年间，是中国古代宫廷建筑之精华，也是中华民族文化的瑰宝之一。故宫占地面积约72万平方米，建筑面积约15万平方米，是世界上现存最大、保存最完整的古代宫殿建筑群之一。

\quad 

故宫建筑群布局严谨、精美，包括乾清宫、交泰殿、保和殿、午门、三大殿、太和殿、乾清宫、养心殿、紫禁城等108个建筑和部门，有9000多间房间，拥有珍贵的文物和艺术品，被誉为世界瑰宝。

\quad 

故宫的建筑风格以黄色琉璃瓦顶、红墙黄瓦、檐角飞檐为主，色彩鲜艳，富丽堂皇。故宫的文物丰富多彩，包括宫廷用品、艺术品、图书、典章制度、历史文献等，被誉为中国古代文化之宝库。

\quad 

故宫是中国历史和文化的重要象征，也是中国人民的珍贵遗产，吸引着众多国内外游客前来参观和探索。
\end{CJK}

(Translated in English:) The Forbidden City, located in the center of Beijing within the Imperial Palace complex, is a representation of the royal palaces of feudal dynasties in China and one of the most complete ancient palace architectural complexes in the world. It is one of the most important cultural heritages in China. The construction of the Forbidden City began in the Yongle reign of the Ming Dynasty, and it represents the essence of ancient Chinese court architecture and one of the treasures of Chinese culture. The Forbidden City covers an area of about 720,000 square meters, with a building area of about 150,000 square meters, making it one of the largest and most complete ancient palace architectural complexes in the world. The layout of the Forbidden City is rigorous and exquisite, including 108 buildings and departments such as the Qianqing Palace, Jiaotai Hall, Baohe Hall, Meridian Gate, Three Great Halls, Taihe Palace, Qianqing Palace, Yangxin Hall, and the Purple Forbidden City. It has more than 9,000 rooms and houses precious cultural relics and artworks, making it a world treasure. The architectural style of the Forbidden City is characterized by yellow glazed tile roofs, red walls, yellow tiles, and flying eaves, with bright colors and magnificent grandeur. The cultural relics in the Forbidden City are rich and diverse, including court articles, artworks, books, systems of laws and regulations, and historical documents, making it a treasure trove of ancient Chinese culture. The Forbidden City is an important symbol of Chinese history and culture, as well as a precious heritage of the Chinese people, attracting numerous domestic and foreign tourists to visit and explore.
\\
\midrule
User & Introduce the place in details \\
LLaVA~\cite{liu2023visual} &  The image features a large and beautiful red palace, which appears to be a Chinese temple or government building, situated on a grassy hill. The main building is surrounded by several smaller buildings, all of which are adorned with red flags. These flags display the Chinese flag design, emphasizing the cultural significance of the location.

In the foreground, there are various vehicles, including a few cars and trucks, parked close to the palace grounds. Additionally, a person can be seen standing further away from the main building, possibly admiring the architecture or waiting for someone. With its grand architecture and serene setting, the palace serves as an impressive landmark in the area.
\\ \midrule
User & Introduce the place in details \\
MiniGPT-4~\cite{zhu2023minigpt4} & The image shows a red building with a large gate in the front. The building has a large red roof and white walls. There are several flags on the roof of the building, including the Chinese flag. The building is surrounded by green grass and bushes. The sky is clear and blue.
\\
\bottomrule
\end{tabular}
}
\vspace{2mm}
\captionof{table}{Example prompt demonstrating \shortname{}, LLaVA and MiniGPT-4's Chinese characteristic visual input capability. The prompt requires image understanding.} 
\label{tab:visual_example_forbidden}  
  \end{minipage}
\end{table}

\begin{table}[t]
  \begin{minipage}{0.99\textwidth}
\centering  
\vspace{-4mm}
 
\scalebox{0.88}{
\begin{tabular}{l p{12.5cm} }
\toprule
 \multicolumn{2}{l}{\bf Visual input example, Honor of Kings:}  \\
\midrule
&  \includegraphics[height=5.5cm]{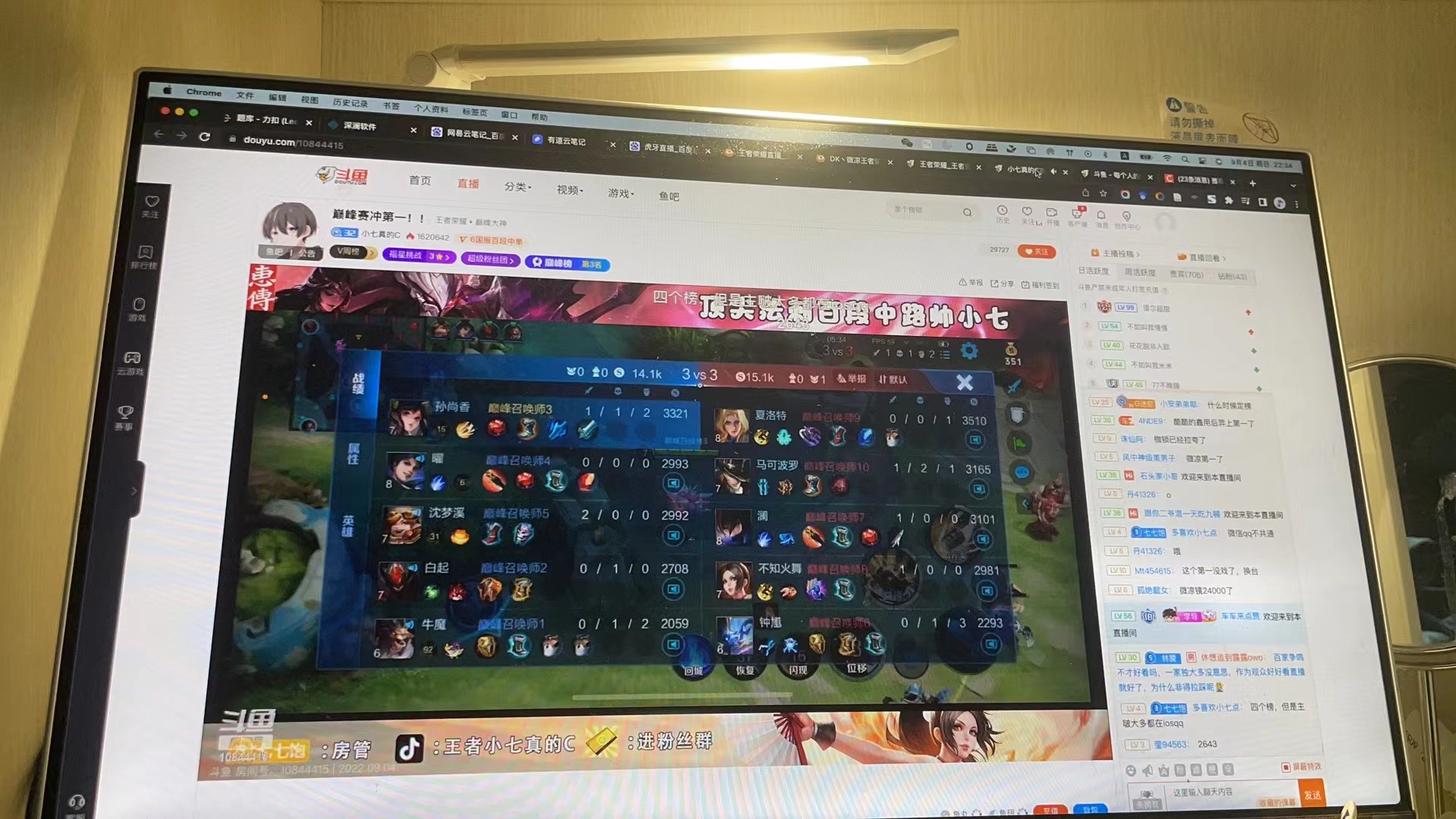} \\
\midrule
User &  \begin{CJK}{UTF8}{gbsn}这是什么游戏？介绍一下。\end{CJK} (What game is this? Introduce to me.)\\
\shortname{} & \begin{CJK}{UTF8}{gbsn}
王者荣耀是由腾讯游戏开发并发行的一款多人在线战斗竞技手游，于2015年11月26日在中国内地正式发行，是一款竞技游戏，玩家可以选择不同的英雄进行战斗，并与其他玩家组队竞技。王者荣耀在中国受到了广泛欢迎，并且在全球范围内也拥有大量的玩家。
\end{CJK}

(Translated in English:) Honor of Kings is a multiplayer online battle arena (MOBA) mobile game developed and published by Tencent Games. It was officially released in mainland China on November 26, 2015. As a competitive game, players can select different heroes to fight and compete with other players in teams. Honor of Kings has gained widespread popularity in China and has a large player base globally.
\\
\midrule
User & What game is this? Introduce to me. \\
LLaVA~\cite{liu2023visual} &  The image shows a computer screen with a game that appears to involve a combination of elements from popular games like snake and pokemon. The game is being played with a mouse, and the screen is filled with many small images, likely representing different characters or items within the game. The game seems to be set in a grassy field, where the player encounters and engages with various creatures or obstacles. As an AI, I am unable to provide further information on the specific game or its rules, but it is clear that the user is actively engaged in playing this unique game on their computer.
\\ \midrule
User & What game is this? Introduce to me. \\
MiniGPT-4~\cite{zhu2023minigpt4} & This is a screenshot of a game with a character selection screen. The game appears to be a multiplayer game with various characters to choose from. The game appears to be a role-playing game with different characters and abilities. 
\\
\bottomrule
\end{tabular}
}
\vspace{2mm}
\captionof{table}{Example prompt demonstrating \shortname{}, LLaVA and MiniGPT-4's Chinese characteristic visual input capability. The prompt requires image understanding.} 

\vspace{-10pt}
\label{tab:visual_example_kings}  
  \end{minipage}
\end{table}

\subsection{ASR and Multimodal ASR}
We evaluate the ASR and multmodal ASR (MASR) on AISHELL-2 and VSDial-CN datasets. Please refer to Appendix~\ref{lab:implementation} for details.

As shown in Table~\ref{tab:asr}, we compare X-LLM with the CIF-based model mentioned in Section~\ref{sec:firststage}. From the results in the table, overall, the speech recognition performance of X-LLM is weaker than that of the CIF-based Model. This is because the CIF-based model is trained with full parameters, while X-LLM only trains the BERT in the C-Former and the S-Adapter. In the results of multimodal speech recognition on the VSDial-CN dataset, the introduction of images caused a decrease in the speech recognition performance of X-LLM (from 3.4 to 4.7 on test-orig and from 4.7 to 6.8 on test-art). This may be because we provided X-LLM with a fixed single instruction ``Please faithfully recognize the speech'' during the second training stage for speech recognition. However, when conducting multimodal speech recognition, we replace the instruction with ``Please faithfully recognize the speech based on the image.''. After finetuning X-LLM with 2k speech recognition-related instructions in the third stage, the error rate of X-LLM's multimodal speech recognition decreased from 4.7 to 4.3 on test-orig and from 6.8 to 6.3 on test-art. In the future, we will supplement the training process by using various speech recognition instructions in the second stage and increasing the amount of data for finetuning instructions in the third stage to observe the changes in the multimodal speech recognition ability of X-LLM. Additionally, a more powerful LLM may have stronger instruction generalization, which could improve the performance of multimodal speech recognition.

We observe that although the addition of images to X-LLM's speech recognition task results in a slight decrease in performance, X-LLM is able to comprehend spoken questions in speech without finetuning, and provide appropriate responses. It can also incorporate images to provide suitable answers to spoken questions. After a small amount of data finetuning in the third phase, X-LLM's ability in this regard is further improved.

\begin{table}[t]
  \centering
 
  \resizebox{0.8\linewidth}{!}{
    \begin{tabular}{l|ccc|cccc}
    \toprule
    \multicolumn{1}{c|}{\multirow{3}[6]{*}{Model}} & \multicolumn{3}{c|}{AISHELL-2} & \multicolumn{4}{c}{VSDial-CN} \\
\cmidrule{2-8}          & \multirow{2}[4]{*}{test-ios} & \multirow{2}[4]{*}{test-android} & \multirow{2}[4]{*}{test-mic} & \multicolumn{2}{c}{test-orig} & \multicolumn{2}{c}{test-art} \\
\cmidrule{5-8}          &       &       &       & S     & S + V & S     & S + V \\
    \midrule
    CIF-based Model &    6.0   &  6.4     &   6.4    &   2.8    &  1.7     &   6.0    &  5.3 \\
    X-LLM &   8.7    &  9.0     &    9.4   &    3.4   &   4.7 / 4.3    &  4.7    &  6.8 / 6.3 \\
    \bottomrule
    \end{tabular}}%
    \vspace{2mm}
     \caption{Performance of X-LLM on AISHELL-2 and VSDial-CN (CER\%). ``S'' denotes only using speech for ASR. ``S+V'' denotes using speech and images for MASR. }
  \label{tab:asr} 
\end{table}%

\section{Discussions}
This paper demonstrates the effectiveness of X-LLM, which injects multiple modalities as foreign languages into a large language model through the X2L interface, endowing LLM with powerful multimodal capabilities. We design a three-stage training method to train X-LLM, where each modality interface has high independence in the first two stages, facilitating simultaneous training. Through the first two stages of training, X-LLM can interact with each modality through language. Furthermore, X-LLM can complete tasks involving multiple modalities (such as visual spoken question answering) without further finetuning on joint instruction datasets, thanks to its integration of the instruction generalization ability of large language models and its adaptation to the multimodal domain. The integration of multiple modalities without training greatly facilitates the modality expansion of X-LLM. To further explore the impact of joint multimodal instruction data on X-LLM's ability to integrate multiple modalities, we construct a streamlined but high-quality multimodal instruction dataset, and X-LLM's performance is further improved after fine-tuning on this data.

This project is still ongoing and currently has several limitations: (1) Limitations of the language model. X-LLM is built on top of ChatGLM with only 6B parameters and inherits its limitations, including but not limited to unreliable reasoning ability and fabrication of non-existent facts.
(2) Insufficient training for modal connections. X-LLM's multi-modal perception ability is somewhat limited. We only used a small amount of multi-modal data sets to connect the multi-modal encoder and a large language model. 
There are several directions for further exploration: (1) Data scale. Compared to BLIP2, we only used a small amount of Chinese multimodal data. We believe that using larger Chinese data for training can significantly improve the model's performance by increasing concept coverage. 
(2) Connecting more modalities. We can connect audio to enable LLM to understand and interact with non-verbal audio. We can also connect the status information of various terminals to LLM, so that LLM can control the terminals based on their status information. (3) Using better LLM. 
Due to the limitation of computing resources, we only used a 6B language model for experimentation. 
It can be expected that using a stronger language model, X-LLM will gain more powerful capabilities.

\bibliography{egbib}

\begin{thebibliography}{10}

\bibitem{antol2015vqa}
Stanislaw Antol, Aishwarya Agrawal, Jiasen Lu, Margaret Mitchell, Dhruv Batra,
  C~Lawrence Zitnick, and Devi Parikh.
\newblock Vqa: Visual question answering.
\newblock In {\em Proceedings of the IEEE international conference on computer
  vision}, pages 2425--2433, 2015.

\bibitem{bi2020palm}
Bin Bi, Chenliang Li, Chen Wu, Ming Yan, Wei Wang, Songfang Huang, Fei Huang,
  and Luo Si.
\newblock Palm: Pre-training an autoencoding\&autoregressive language model for
  context-conditioned generation.
\newblock {\em arXiv preprint arXiv:2004.07159}, 2020.

\bibitem{bu2017aishell}
Hui Bu, Jiayu Du, Xingyu Na, Bengu Wu, and Hao Zheng.
\newblock Aishell-1: An open-source mandarin speech corpus and a speech
  recognition baseline.
\newblock In {\em 2017 20th conference of the oriental chapter of the
  international coordinating committee on speech databases and speech I/O
  systems and assessment (O-COCOSDA)}, pages 1--5. IEEE, 2017.

\bibitem{chen2022utc}
Cheng Chen, Zhenshan Tan, Qingrong Cheng, Xin Jiang, Qun Liu, Yudong Zhu, and
  Xiaodong Gu.
\newblock Utc: a unified transformer with inter-task contrastive learning for
  visual dialog.
\newblock In {\em Proceedings of the IEEE/CVF Conference on computer vision and
  pattern recognition}, pages 18103--18112, 2022.

\bibitem{chen2023vlp}
Fei-Long Chen, Du-Zhen Zhang, Ming-Lun Han, Xiu-Yi Chen, Jing Shi, Shuang Xu,
  and Bo~Xu.
\newblock Vlp: A survey on vision-language pre-training.
\newblock {\em Machine Intelligence Research}, 20(1):38--56, 2023.

\bibitem{chen2020dmrm}
Feilong Chen, Fandong Meng, Jiaming Xu, Peng Li, Bo~Xu, and Jie Zhou.
\newblock Dmrm: A dual-channel multi-hop reasoning model for visual dialog.
\newblock In {\em Proceedings of the AAAI Conference on Artificial
  Intelligence}, volume~34, pages 7504--7511, 2020.

\bibitem{chen2022unsupervised}
Feilong Chen, Duzhen Zhang, Xiuyi Chen, Jing Shi, Shuang Xu, and Bo~Xu.
\newblock Unsupervised and pseudo-supervised vision-language alignment in
  visual dialog.
\newblock In {\em Proceedings of the 30th ACM International Conference on
  Multimedia}, pages 4142--4153, 2022.

\bibitem{chiang2023vicuna}
Wei-Lin Chiang, Zhuohan Li, Zi~Lin, Ying Sheng, Zhanghao Wu, Hao Zhang, Lianmin
  Zheng, Siyuan Zhuang, Yonghao Zhuang, Joseph~E Gonzalez, et~al.
\newblock Vicuna: An open-source chatbot impressing gpt-4 with 90 quality,
  2023.

\bibitem{das2017visual}
Abhishek Das, Satwik Kottur, Khushi Gupta, Avi Singh, Deshraj Yadav,
  Jos{\'e}~MF Moura, Devi Parikh, and Dhruv Batra.
\newblock Visual dialog.
\newblock In {\em Proceedings of the IEEE conference on computer vision and
  pattern recognition}, pages 326--335, 2017.

\bibitem{dehghani2023scaling}
Mostafa Dehghani, Josip Djolonga, Basil Mustafa, Piotr Padlewski, Jonathan
  Heek, Justin Gilmer, Andreas Steiner, Mathilde Caron, Robert Geirhos, Ibrahim
  Alabdulmohsin, et~al.
\newblock Scaling vision transformers to 22 billion parameters.
\newblock {\em arXiv preprint arXiv:2302.05442}, 2023.

\bibitem{devlin2018bert}
Jacob Devlin, Ming-Wei Chang, Kenton Lee, and Kristina Toutanova.
\newblock Bert: Pre-training of deep bidirectional transformers for language
  understanding.
\newblock {\em arXiv preprint arXiv:1810.04805}, 2018.

\bibitem{Dong020}
Linhao Dong and Bo~Xu.
\newblock {CIF:} continuous integrate-and-fire for end-to-end speech
  recognition.
\newblock In {\em {ICASSP}}, pages 6079--6083. {IEEE}, 2020.

\bibitem{dosovitskiy2020image}
Alexey Dosovitskiy, Lucas Beyer, Alexander Kolesnikov, Dirk Weissenborn,
  Xiaohua Zhai, Thomas Unterthiner, Mostafa Dehghani, Matthias Minderer, Georg
  Heigold, Sylvain Gelly, et~al.
\newblock An image is worth 16x16 words: Transformers for image recognition at
  scale.
\newblock {\em arXiv preprint arXiv:2010.11929}, 2020.

\bibitem{driess2023palm}
Danny Driess, Fei Xia, Mehdi~SM Sajjadi, Corey Lynch, Aakanksha Chowdhery,
  Brian Ichter, Ayzaan Wahid, Jonathan Tompson, Quan Vuong, Tianhe Yu, et~al.
\newblock Palm-e: An embodied multimodal language model.
\newblock {\em arXiv preprint arXiv:2303.03378}, 2023.

\bibitem{du2018aishell}
Jiayu Du, Xingyu Na, Xuechen Liu, and Hui Bu.
\newblock Aishell-2: Transforming mandarin asr research into industrial scale.
\newblock {\em arXiv preprint arXiv:1808.10583}, 2018.

\bibitem{du2022survey}
Yifan Du, Zikang Liu, Junyi Li, and Wayne~Xin Zhao.
\newblock A survey of vision-language pre-trained models.
\newblock {\em arXiv preprint arXiv:2202.10936}, 2022.

\bibitem{du2022glm}
Zhengxiao Du, Yujie Qian, Xiao Liu, Ming Ding, Jiezhong Qiu, Zhilin Yang, and
  Jie Tang.
\newblock Glm: General language model pretraining with autoregressive blank
  infilling.
\newblock In {\em Proceedings of the 60th Annual Meeting of the Association for
  Computational Linguistics (Volume 1: Long Papers)}, pages 320--335, 2022.

\bibitem{gao2017video}
Lianli Gao, Zhao Guo, Hanwang Zhang, Xing Xu, and Heng~Tao Shen.
\newblock Video captioning with attention-based lstm and semantic consistency.
\newblock {\em IEEE Transactions on Multimedia}, 19(9):2045--2055, 2017.

\bibitem{girshick2015fast}
Ross Girshick.
\newblock Fast r-cnn.
\newblock In {\em Proceedings of the IEEE international conference on computer
  vision}, pages 1440--1448, 2015.

\bibitem{gu2022wukong}
Jiaxi Gu, Xiaojun Meng, Guansong Lu, Lu~Hou, Niu Minzhe, Xiaodan Liang, Lewei
  Yao, Runhui Huang, Wei Zhang, Xin Jiang, et~al.
\newblock Wukong: A 100 million large-scale chinese cross-modal pre-training
  benchmark.
\newblock {\em Advances in Neural Information Processing Systems},
  35:26418--26431, 2022.

\bibitem{gulati2020conformer}
Anmol Gulati, James Qin, Chung{-}Cheng Chiu, Niki Parmar, Yu~Zhang, Jiahui Yu,
  Wei Han, Shibo Wang, Zhengdong Zhang, Yonghui Wu, and Ruoming Pang.
\newblock Conformer: Convolution-augmented transformer for speech recognition.
\newblock In {\em {INTERSPEECH}}, pages 5036--5040. {ISCA}, 2020.

\bibitem{HanDLCZMX22}
Minglun Han, Linhao Dong, Zhenlin Liang, Meng Cai, Shiyu Zhou, Zejun Ma, and
  Bo~Xu.
\newblock Improving end-to-end contextual speech recognition with fine-grained
  contextual knowledge selection.
\newblock In {\em {ICASSP}}, pages 8532--8536. {IEEE}, 2022.

\bibitem{HanDZX21}
Minglun Han, Linhao Dong, Shiyu Zhou, and Bo~Xu.
\newblock Cif-based collaborative decoding for end-to-end contextual speech
  recognition.
\newblock In {\em {ICASSP}}, pages 6528--6532. {IEEE}, 2021.

\bibitem{huang2023language}
Shaohan Huang, Li~Dong, Wenhui Wang, Yaru Hao, Saksham Singhal, Shuming Ma,
  Tengchao Lv, Lei Cui, Owais~Khan Mohammed, Qiang Liu, et~al.
\newblock Language is not all you need: Aligning perception with language
  models.
\newblock {\em arXiv preprint arXiv:2302.14045}, 2023.

\bibitem{khan2022transformers}
Salman Khan, Muzammal Naseer, Munawar Hayat, Syed~Waqas Zamir, Fahad~Shahbaz
  Khan, and Mubarak Shah.
\newblock Transformers in vision: A survey.
\newblock {\em ACM computing surveys (CSUR)}, 54(10s):1--41, 2022.

\bibitem{kim2021vilt}
Wonjae Kim, Bokyung Son, and Ildoo Kim.
\newblock Vilt: Vision-and-language transformer without convolution or region
  supervision.
\newblock In {\em International Conference on Machine Learning}, pages
  5583--5594. PMLR, 2021.

\bibitem{krishna2017dense}
Ranjay Krishna, Kenji Hata, Frederic Ren, Li~Fei-Fei, and Juan Carlos~Niebles.
\newblock Dense-captioning events in videos.
\newblock In {\em Proceedings of the IEEE international conference on computer
  vision}, pages 706--715, 2017.

\bibitem{krishna2017visual}
Ranjay {Krishna}, Yuke {Zhu}, Oliver {Groth}, Justin {Johnson}, Kenji {Hata},
  Joshua {Kravitz}, Stephanie {Chen}, Yannis {Kalantidis}, Li-Jia {Li},
  David~A. {Shamma}, Michael~S. {Bernstein}, and Li~{Fei-Fei}.
\newblock Visual genome: Connecting language and vision using crowdsourced
  dense image annotations.
\newblock {\em International Journal of Computer Vision}, 123(1):32--73, 2017.

\bibitem{li2023blip2}
Junnan Li, Dongxu Li, Silvio Savarese, and Steven Hoi.
\newblock Blip-2: Bootstrapping language-image pre-training with frozen image
  encoders and large language models.
\newblock {\em arXiv preprint arXiv:2301.12597}, 2023.

\bibitem{li2022blip}
Junnan Li, Dongxu Li, Caiming Xiong, and Steven Hoi.
\newblock Blip: Bootstrapping language-image pre-training for unified
  vision-language understanding and generation.
\newblock In {\em International Conference on Machine Learning}, pages
  12888--12900. PMLR, 2022.

\bibitem{li2019visualbert}
Liunian~Harold Li, Mark Yatskar, Da~Yin, Cho-Jui Hsieh, and Kai-Wei Chang.
\newblock Visualbert: A simple and performant baseline for vision and language.
\newblock {\em arXiv preprint arXiv:1908.03557}, 2019.

\bibitem{lin2014microsoft}
Tsung-Yi Lin, Michael Maire, Serge Belongie, James Hays, Pietro Perona, Deva
  Ramanan, Piotr Doll{\'a}r, and C~Lawrence Zitnick.
\newblock Microsoft coco: Common objects in context.
\newblock In {\em Proceedings of the European Conference on Computer Vision},
  pages 740--755. Springer, 2014.

\bibitem{liu2023visual}
Haotian Liu, Chunyuan Li, Qingyang Wu, and Yong~Jae Lee.
\newblock Visual instruction tuning.
\newblock {\em arXiv preprint arXiv:2304.08485}, 2023.

\bibitem{long2022vision}
Siqu Long, Feiqi Cao, Soyeon~Caren Han, and Haiqing Yang.
\newblock Vision-and-language pretrained models: A survey.
\newblock {\em arXiv preprint arXiv:2204.07356}, 2022.

\bibitem{lu2019vilbert}
Jiasen Lu, Dhruv Batra, Devi Parikh, and Stefan Lee.
\newblock Vilbert: Pretraining task-agnostic visiolinguistic representations
  for vision-and-language tasks.
\newblock {\em Advances in neural information processing systems}, 32, 2019.

\bibitem{murahari2020large}
Vishvak Murahari, Dhruv Batra, Devi Parikh, and Abhishek Das.
\newblock Large-scale pretraining for visual dialog: A simple state-of-the-art
  baseline.
\newblock In {\em Computer Vision--ECCV 2020: 16th European Conference,
  Glasgow, UK, August 23--28, 2020, Proceedings, Part XVIII}, pages 336--352.
  Springer, 2020.

\bibitem{openaigpt4}
OpenAI.
\newblock Gpt-4 technical report, 2023.

\bibitem{ordonez2011im2text}
Vicente Ordonez, Girish Kulkarni, and Tamara~L Berg.
\newblock Im2text: Describing images using 1 million captioned photographs.
\newblock In {\em Advances in Neural Information Processing Systems}, pages
  1143--1151, 2011.

\bibitem{Park2019SpecAugmentAS}
Daniel~S. Park, William Chan, Yu~Zhang, Chung-Cheng Chiu, Barret Zoph,
  Ekin~Dogus Cubuk, and Quoc~V. Le.
\newblock Specaugment: A simple data augmentation method for automatic speech
  recognition.
\newblock In {\em Interspeech}, 2019.

\bibitem{plummer2015flickr30k}
Bryan~A Plummer, Liwei Wang, Chris~M Cervantes, Juan~C Caicedo, Julia
  Hockenmaier, and Svetlana Lazebnik.
\newblock Flickr30k entities: Collecting region-to-phrase correspondences for
  richer image-to-sentence models.
\newblock In {\em Proceedings of the IEEE international conference on computer
  vision}, pages 2641--2649, 2015.

\bibitem{radford2021learning}
Alec Radford, Jong~Wook Kim, Chris Hallacy, Aditya Ramesh, Gabriel Goh,
  Sandhini Agarwal, Girish Sastry, Amanda Askell, Pamela Mishkin, Jack Clark,
  et~al.
\newblock Learning transferable visual models from natural language
  supervision.
\newblock In {\em International conference on machine learning}, pages
  8748--8763. PMLR, 2021.

\bibitem{radford2019language}
Alec Radford, Jeffrey Wu, Rewon Child, David Luan, Dario Amodei, Ilya
  Sutskever, et~al.
\newblock Language models are unsupervised multitask learners.
\newblock {\em OpenAI blog}, 1(8):9, 2019.

\bibitem{raffel2020exploring}
Colin Raffel, Noam Shazeer, Adam Roberts, Katherine Lee, Sharan Narang, Michael
  Matena, Yanqi Zhou, Wei Li, and Peter~J Liu.
\newblock Exploring the limits of transfer learning with a unified text-to-text
  transformer.
\newblock {\em The Journal of Machine Learning Research}, 21(1):5485--5551,
  2020.

\bibitem{0006H0QZZL21}
Yi~Ren, Chenxu Hu, et~al.
\newblock Fastspeech 2: Fast and high-quality end-to-end text to speech.
\newblock In {\em Proc. {ICLR}}, 2021.

\bibitem{sharma2018conceptual}
Piyush Sharma, Nan Ding, Sebastian Goodman, and Radu Soricut.
\newblock Conceptual captions: A cleaned, hypernymed, image alt-text dataset
  for automatic image captioning.
\newblock In {\em Proceedings of the 56th Annual Meeting of the Association for
  Computational Linguistics (Volume 1: Long Papers)}, pages 2556--2565, 2018.

\bibitem{shen2023hugginggpt}
Yongliang Shen, Kaitao Song, Xu~Tan, Dongsheng Li, Weiming Lu, and Yueting
  Zhuang.
\newblock Hugginggpt: Solving ai tasks with chatgpt and its friends in
  huggingface.
\newblock {\em arXiv preprint arXiv:2303.17580}, 2023.

\bibitem{shi2020aishell}
Yao Shi, Hui Bu, et~al.
\newblock Aishell-3: A multi-speaker mandarin tts corpus and the baselines.
\newblock {\em arXiv preprint arXiv:2010.11567}, 2020.

\bibitem{touvron2023llama}
Hugo Touvron, Thibaut Lavril, Gautier Izacard, Xavier Martinet, Marie-Anne
  Lachaux, Timoth{\'e}e Lacroix, Baptiste Rozi{\`e}re, Naman Goyal, Eric
  Hambro, Faisal Azhar, et~al.
\newblock Llama: Open and efficient foundation language models.
\newblock {\em arXiv preprint arXiv:2302.13971}, 2023.

\bibitem{vaswani2017attention}
Ashish Vaswani, Noam Shazeer, Niki Parmar, Jakob Uszkoreit, Llion Jones,
  Aidan~N Gomez, {\L}ukasz Kaiser, and Illia Polosukhin.
\newblock Attention is all you need.
\newblock {\em Advances in neural information processing systems}, 30, 2017.

\bibitem{vinyals2016show}
Oriol Vinyals, Alexander Toshev, Samy Bengio, and Dumitru Erhan.
\newblock Show and tell: Lessons learned from the 2015 mscoco image captioning
  challenge.
\newblock {\em IEEE transactions on pattern analysis and machine intelligence},
  39(4):652--663, 2016.

\bibitem{wang2020vd}
Yue Wang, Shafiq Joty, Michael Lyu, Irwin King, Caiming Xiong, and Steven~CH
  Hoi.
\newblock Vd-bert: A unified vision and dialog transformer with bert.
\newblock In {\em Proceedings of the 2020 Conference on Empirical Methods in
  Natural Language Processing (EMNLP)}, pages 3325--3338, 2020.

\bibitem{wen2015semantically}
Tsung-Hsien Wen, Milica Gasic, Nikola Mrksic, Pei-Hao Su, David Vandyke, and
  Steve Young.
\newblock Semantically conditioned lstm-based natural language generation for
  spoken dialogue systems.
\newblock {\em arXiv preprint arXiv:1508.01745}, 2015.

\bibitem{wu2023visual}
Chenfei Wu, Shengming Yin, Weizhen Qi, Xiaodong Wang, Zecheng Tang, and Nan
  Duan.
\newblock Visual chatgpt: Talking, drawing and editing with visual foundation
  models.
\newblock {\em arXiv preprint arXiv:2303.04671}, 2023.

\bibitem{wu2017ai}
Jiahong Wu, He~Zheng, Bo~Zhao, Yixin Li, Baoming Yan, Rui Liang, Wenjia Wang,
  Shipei Zhou, Guosen Lin, Yanwei Fu, et~al.
\newblock Ai challenger: A large-scale dataset for going deeper in image
  understanding.
\newblock {\em arXiv preprint arXiv:1711.06475}, 2017.

\bibitem{Xu_2016_CVPR}
Jun Xu, Tao Mei, Ting Yao, and Yong Rui.
\newblock Msr-vtt: A large video description dataset for bridging video and
  language.
\newblock In {\em Proceedings of the IEEE Conference on Computer Vision and
  Pattern Recognition (CVPR)}, June 2016.

\bibitem{zeng2023glm-130b}
Aohan Zeng, Xiao Liu, Zhengxiao Du, Zihan Wang, Hanyu Lai, Ming Ding, Zhuoyi
  Yang, Yifan Xu, Wendi Zheng, Xiao Xia, Weng~Lam Tam, Zixuan Ma, Yufei Xue,
  Jidong Zhai, Wenguang Chen, Zhiyuan Liu, Peng Zhang, Yuxiao Dong, and Jie
  Tang.
\newblock {GLM}-130b: An open bilingual pre-trained model.
\newblock In {\em The Eleventh International Conference on Learning
  Representations (ICLR)}, 2023.

\bibitem{zhai2022scaling}
Xiaohua Zhai, Alexander Kolesnikov, Neil Houlsby, and Lucas Beyer.
\newblock Scaling vision transformers.
\newblock In {\em Proceedings of the IEEE/CVF Conference on Computer Vision and
  Pattern Recognition}, pages 12104--12113, 2022.

\bibitem{zhou2020unified}
Luowei Zhou, Hamid Palangi, Lei Zhang, Houdong Hu, Jason Corso, and Jianfeng
  Gao.
\newblock Unified vision-language pre-training for image captioning and vqa.
\newblock In {\em Proceedings of the AAAI conference on artificial
  intelligence}, volume~34, pages 13041--13049, 2020.

\bibitem{zhu2023minigpt4}
Deyao Zhu, Jun Chen, Xiaoqian Shen, Xiang Li, and Mohamed Elhoseiny.
\newblock Minigpt-4: Enhancing vision-language understanding with advanced
  large language models.
\newblock {\em arXiv preprint arXiv:2304.10592}, 2023.

\bibitem{zoph2020rethinking}
Barret Zoph, Golnaz Ghiasi, Tsung-Yi Lin, Yin Cui, Hanxiao Liu, Ekin~Dogus
  Cubuk, and Quoc Le.
\newblock Rethinking pre-training and self-training.
\newblock {\em Advances in neural information processing systems},
  33:3833--3845, 2020.

\end{thebibliography}
\bibliographystyle{plain}

\clearpage
\appendix
\section{Implementation Details} \label{lab:implementation}
\subsection{Training Data}

\begin{table}[h]
    \centering
    \resizebox{0.99\columnwidth}!{
    \begin{tabular}{lcccccc}
    \toprule
     Type & Training Stage & Training Module & Source & \#Sample \\
     \midrule
     Image-Text &  The first two stages & Image Interface &  \makecell{CC3M, COCO, VG-Caps, \\Flickr30k, SBU, AI-Caps, Wukong} & 14M \\
     \midrule
     Video-Text &  The second stage & Video Interface &  MSRVTT & 7.7k \\
     \midrule
    Speech-Text  & The first stage & Speech Encoder, CIF module in C-Former & AISHELL-1, AISHELL-2, VSDial-CN & 128k + 1M + 1.2M \\
    Speech-Text  & The second stage & Speech Interface &  AISHELL-2, VSDial-CN & 1M + 370K \\
     \midrule
     Multimodal Instruction & The third stage & I-Adapter, V-Adapter, S-Adapter & \makecell{MiniGPT-4, AISHELL-2, \\ VSDial-CN, ActivityNet Caps} & 10k \\
    \bottomrule
    \end{tabular}}
    \vspace{2mm}
     \caption{Statistics on the datasets of three-stage training.}
    \label{tab:data}
    \vspace{-10pt}
\end{table}

We construct training datasets by incorporating vision-language data, video-language data and automatic ASR (ASR) data and visual spoken dialogue data. The statistics on the training datasets are listed in Table~\ref{tab:data}.

\paragraph{Data for the first two training stages.}
For vision-language data, we mainly apply image-tet pairs, including image-caption pairs, a few video-caption pairs. For the first two training stages of the image interface, we collect Conceptual Caption 3M (CC3M)~\cite{sharma2018conceptual}, MSCOCO image captions (COCO)~\cite{lin2014microsoft}, Visual Genome Captions (VG Captions)~\cite{krishna2017visual}, Flickr30k~\cite{plummer2015flickr30k}, SBU caption (SBU)~\cite{ordonez2011im2text}, AI Challenger captions (AI-Caps)~\cite{wu2017ai}, Wukong Captions (Wukong)~\cite{gu2022wukong}. Please note that except AI Challenger Caption and Wukong is Chinese data, other datasets are English. We use machine translation to translate them into Chinese. For the second training stage of the video interface, we translate MSRVTT datasets~\cite{Xu_2016_CVPR} into Chinese and use the training split to train the video interface. Note that the original  Wukong dataset consists of approximately 100 million image pairs. We use pretrained CLIP~\cite{radford2021learning} to filter out 10M pairs with a similarity greater than 0.45 as training data. We use all image-text data except for the Wukong dataset to form a 4M training set. We use only 4M without using pretrained Q-Former from BLIP2 to train the image interface, we find that the performance of the model decreases due to its inability to relate some of the knowledge behind the images, such as the inability to recognize the movie Titanic.

For the training of the CIF-based ASR model in the first training stage, we use AISHELL-1~\cite{bu2017aishell}, AISHELL-2~\cite{du2018aishell} and a self-constructed multi-modal ASR dataset called VSDial-CN. For the second training stage of the speech interface, we use AISHELL-2~\cite{du2018aishell} and the VSDial-CN dataset as training datasets.

AISHELL-2~\cite{du2018aishell} is an open-source Mandarin Chinese ASR dataset that contains approximately 1,000 hours of speech data. The speech utterance contains 12 domains, such as keywords, voice command, smart home, autonomous driving, and industrial production. 1991 speakers from different accent areas in China participated in this recording. The manual transcription accuracy rate is above 96\%, through professional speech annotation and strict quality inspection. AISHELL-2 consists of three test splits, test-ios, test-android and test-mic whose speech are recoded from iPhones, Android phones, and microphone, respectively.

VSDial-CN (Visual Spoken Dialogue for Chinese, VSDial-CN) is constructed from the visual dialogue (VisDial) dataset~\cite{das2017visual}, which consists of around 120,000 images, each with a caption and 10 rounds of dialogue comprising a question and an answer. The dialogue revolves closely around the given image and the caption, so there is a strong correlation between the image and each round of the dialogue. In the context of multimodal ASR, we use the image as the visual input and the caption as the linguistic input to help recognize the speech utterance of the question in each round of dialogue. Therefore, the VSDial-CN dataset has approximately 1.2 million multimodal ASR training samples. Due to overlapping questions, the number of unique speech utterances of questions after deduplication is approximately 370,000. For VSDial-CN, we first translated all the text in the VisDial dataset into Chinese with a public translation model\footnote{\url{https://huggingface.co/facebook/mbart-large-50-many-to-many-mmt}}, and then synthesized all questions into speech with the FastSpeech2 model~\cite{0006H0QZZL21} trained on AISHELL-3~\cite{shi2020aishell}. Next, we extracted 2,000 question-answer pairs from the VisDial test set as a VSDial test set called \texttt{test-orig}. Additionally, according to the questioning style of the VisDial dataset, we artificially generated 300 questions based on images from the VisDial test set to create a VSDial test set called \texttt{test-art}. Finally, we recorded the speech utterances of all questions in test sets using Android phones or iPhones and re-sampled all speech recordings to 16kHz.

\paragraph{Data for the third training stages.}
As described in Section~\ref{instructiondata}, we use ChatGPT to translate 3.5K image-text instructions built by MiniGPT-4~\cite{zhu2023minigpt4}. Then, we manually select 2k data from AISHELL-2~\cite{du2018aishell} and write 5 different instructions for ASR tasks.  We use ChatGPT to translate the ActivityNet dataset~\cite{krishna2017dense}, followed by manually selecting 1k data and writing 5 different instructions for corresponding video-text tasks. We manually select and rewrite 1k data from self-constructed VSDial-CN data, aiming to enable the model to perform dialogue generation tasks based on images and speech. More details of the data can be found in the appendix, including the details of the training data for the first two stages and the multimodal instruction data.

\subsection{Examples of Test Questions}\label{lab:ques}
We use the test set provided by LLaVA, which consists of 30 randomly sampled images
from COCO Val 2014. Each image is associated with one short question, one detailed question, and one complex reasoning question, resulting in a total of 90 questions. We list some questions as shown in Table~\ref{tab:ques}.

\begin{table}[t]
\centering
\scalebox{0.87}{
\begin{tabular}{l|l}
\toprule
Type & Question  \\
\midrule
Conversation  & What is the color of the two suitcases in the image?  \\
Conversation & What are the main objects on the table in the image?  \\
Conversation & How many doughnuts are in the box?  \\
Conversation & What kind of objects are included in the art installation? \\
\midrule
Detail description & Write a detailed description of the given image. \\
Detail description & Explain the visual content of the image in great detail. \\
Detail description & Can you describe the main features of this image for me? \\
Detail description & What's happening in the scene? \\
\midrule
Complex reasoning  & How might the art installation encourage interaction among people? \\
Complex reasoning  & What factors may have led the cat to choose this location for sleep? \\
Complex reasoning  & What might be the reason for the car to be parked on the side of the road? \\
Complex reasoning  & Why might these giraffes be gathering near the same tree? \\
\bottomrule
\end{tabular}
}
\vspace{2mm}
\caption{Examples of test questions.}
\label{tab:ques} 
\vspace{-15pt}
\end{table}

\subsection{Training Details}

\begin{table}[t]
  \centering
  \resizebox{0.99\columnwidth}!{
    \begin{tabular}{lccccccc}
    \toprule
    Training Stage & Training Module & Init. LR/Min. LR/Warmup LR & Warmup Steps & \#Epoch & Batch Size & GPUs & Training Time \\
    \midrule
    First stage & CIF-based ASR Model & - & - &   84  &  $\sim$ 128  &   8    &  3 days   \\
    \midrule
    \multirow{3}[0]{*}{Seoncd stage} & Image Interface &   3e-5/1e-8/1e-6    &   6000    &   5          &   4         &   8    &  6 days \\
          & Video Interface &   1e-5/1e-8/1e-6    &   200    &   5          &   2         &   2    &  1 hours       \\
          & Speech Interface &   1e-4/1e-8/1e-6    &   6000    &   30          &   4         &   8    &  3 days     \\
    \midrule
    Third stage &  \makecell{I-Adapter\\V-Adapter\\S-Adapter}     &    1e-6/1e-6/1e-6   &   0   &   2          &   2        &   2   &  1 hours \\
    \bottomrule
    \end{tabular}
    }%
    \vspace{2mm}
  \caption{Training details of three-stage training. The image interface and video interface of our final model are initialized from pretrained Q-Former of BLIP2 and we omit their first training stage. }
  \label{tab:trainingdetails}%
  \vspace{-15pt}
\end{table}%

X-LLM has 7.6 billion parameters, including all the single-modality encoders and the large language model ChatGLM. The image and video encoders share a pre-trained ViT-g, while the speech encoder consists of convolution layers and a conformer structure. The Q-Formers in the X2L interfaces are implemented with bert-base. The Q-Former uses the parameters initialized from BLIP2's second stage of Q-Former, while the transformer structure in C-Former is initialized with the weights of pre-trained bert-base-chinese\footnote{\url{https://huggingface.co/bert-base-chinese}}. All adapters are linear layers. The length $L_i$ of the quasi-linguistic embeddings for image features is set to 32. We use the AdamW optimizer with $\beta_1  = 0.9$, $\beta_2  = 0.98$, and a weight decay of 0.05. We use cosine learning rate decay. We use images of size 224 x 224, augmented with random resized cropping and horizontal flipping. The number of the gradient accumulation step is 16. All experiments use up to 8 A100-40G GPUs. More details are shown in Table~\ref{tab:trainingdetails}.

\paragraph{The training of the CIF-based ASR model.} For the training of the CIF-based model, we use different settings. We use a learning rate schedule in which we warm up, hold, then decay the learning rate~\cite{Park2019SpecAugmentAS}. The duration of the warm-up stage, hold stage, and decay stage are 24,000 steps, 36,000 steps, and 120,000 steps, respectively. The learning rate of the hold stage is 3e-4. The optimizer has a weight decay of 0.01. The CIF-based ASR model is trained on 8 A100-80G GPUs.

\paragraph{The structure of the CIF-based ASR model.} The speech encoder of the CIF-based ASR model consists of a convolution front-end and a conformer structure~\cite{gulati2020conformer}. The convolution front-end is a 2-dimensional convolution layer with 256 output channels, kernel size 3, and stride 2. The conformer module consists of 18 blocks with $d_{model}=512$, $d_{ffn}=2048$, $h=8$ and kernel size 31 (for depth-wise convolution), and 2 max-pooling layers after the 6th and the 12th blocks. The CIF module contains a 1-dimensional convolution layer with 512 output channels, kernel size 5 and stride 1, and a fully-connected layer followed by a sigmoid activation. The decoder consists of several fully-connected layers and a transformer module \cite{vaswani2017attention} comprised of 6 blocks with $d_{model}=512$, $d_{ffn}=2048$ and $h=8$. To support multimodal inputs in the context of multimodal ASR, the 3rd and the 4th blocks incorporate visual input features via the cross-attention mechanism, and the 5th and the 6th blocks incorporate the linguistic input features via the cross-attention mechanism. The visual inputs are encoded by ViT\footnote{\url{https://huggingface.co/openai/clip-vit-base-patch16}}, and the linguistic inputs are encoded by pre-trained bert-base-chinese. Note that visual and linguistic input features are set to zero vectors when there are no multimodal inputs.

\section{Qualitative Examples}\label{lab:examples}
We provide more examples of video inputs, speech inputs and multimodal inputs. From the examples given, it can be seen that X-LLM has impressive video and speech understanding capabilities. As shown in Table~\ref{tab:video_example1} and \ref{tab:mmt_example1}, X-LLM is able to provide responses based on video content and perform multimodal machine translation, such as translating ``bank'' into Chinese as ``sloping raised land'' rather than ``a financial organization'' according to the given image. Additionally, X-LLM can integrate multiple modalities without the need for further instruction finetuning, such as being able to answer spoken questions based on images without requiring ASR as shown in Table~\ref{tab:vsqa_example}.

In addition, we propose two examples of ethical, moral, and legal design, and X-LLM is able to handle these situations well. In the example shown in Table~\ref{tab:law_example1}, X-LLM refuses to answer the user's question because the image violated relevant Chinese laws, and reminded the user to comply with the relevant laws. In the example shown in Table~\ref{tab:ethics_example1}, X-LLM refuses to make judgments or recommendations on ethical and moral issues and reminded the user to comply with traffic regulations and safety knowledge, drive carefully, and pay attention to safety.

\begin{table}
  \begin{minipage}{0.99\textwidth}
\centering  
\vspace{-4mm}
\scalebox{0.88}{
\begin{tabular}{l p{12.5cm} }
\toprule
 \multicolumn{2}{l}{\bf Video input example, jellyfish}  \\
\midrule
&  \includegraphics[height=5.5cm]{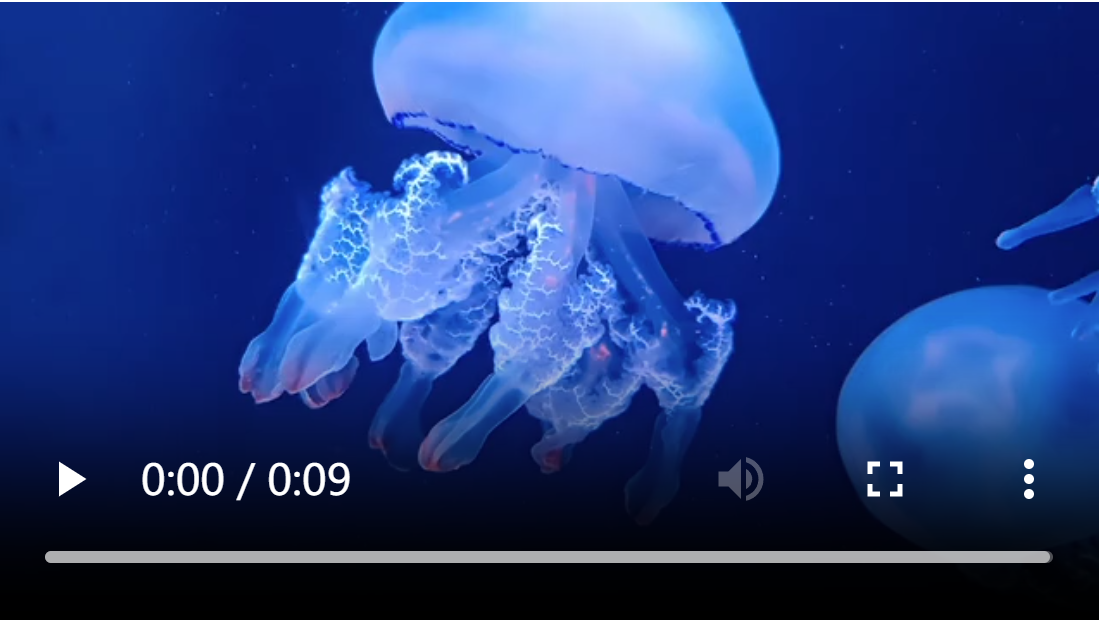} \\
\midrule
User &  \begin{CJK}{UTF8}{gbsn}给这段视频写一个标题。\end{CJK} (Write a titel for this video.)\\
\shortname{} & \begin{CJK}{UTF8}{gbsn}
一只水母在游动
\end{CJK}

(Translated in English:) A jellyfish is swimming
\\
\midrule
User &  \begin{CJK}{UTF8}{gbsn}给这段视频配一段文字，100字左右。\end{CJK} (Write a paragraph of text for this video, approximately 100 words.)\\
\shortname{} & \begin{CJK}{UTF8}{gbsn}
视频中展示了一只水母在蓝色的海洋中游动。水母是一种无脊椎动物，生活在冷水中，是许多动物的食物，但同时也是一类受到过度捕捞的威胁的生物。水母可以通过繁殖来保护种群数量，但过度繁殖也会导致它们无法生存，希望这个视频能够引起人们对海洋生物保护的关注。
\end{CJK} 

(Translated in English:) The video shows a jellyfish swimming in the blue ocean. Jellyfish are invertebrates that live in cold water and are a food source for many animals. However, they are also threatened by overfishing. Jellyfish can protect their populations by reproducing, but over-reproduction can also lead to their inability to survive. Hopefully, this video will raise awareness of the need to protect marine life.
\\
\bottomrule
\end{tabular}
}
\vspace{2mm}
\captionof{table}{Example prompt demonstrating \shortname{}'s video input capability. The prompt requires video understanding.}  
\label{tab:video_example1}  
  \end{minipage}
\end{table}

\begin{table}
  \begin{minipage}{0.99\textwidth}
\centering  
\vspace{-4mm}
\scalebox{0.88}{
\begin{tabular}{l p{12.5cm} }
\toprule
 \multicolumn{2}{l}{\bf Visual spoken question answering example}  \\
\midrule
&  \includegraphics[height=5.5cm]{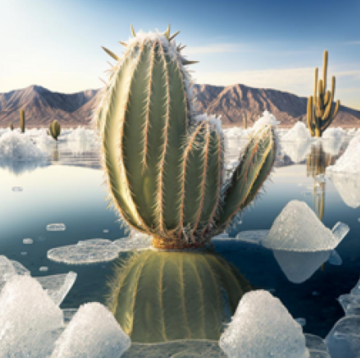} \\
\midrule
User &  (Input a voice) \begin{CJK}{UTF8}{gbsn}详细描述一下这个照片\end{CJK} (Describe this photo in detail)\\
\shortname{} & \begin{CJK}{UTF8}{gbsn}
这是一张由摄影师拍摄的照片，展示了一个被冰覆盖的沙漠和仙人掌的景象。这个沙漠位于南美洲的安第斯山脉附近，被冰雪覆盖着。在沙漠的周围，有一些巨大的仙人掌，它们被冰雪所覆盖，但仍然能够生存。

照片中的仙人掌看起来非常巨大，它们的叶子和茎都被冰雪所覆盖，看起来非常壮观。在沙漠中，还有一些小型的仙人掌，它们生长在冰雪的附近，可能是因为它们需要一些水分来维持生命。

整个景象看起来非常壮观，充满了神秘感。由于该地区的气温非常低，所以冰雪覆盖的沙漠和仙人掌非常特别，是摄影师非常喜欢拍摄的主题之一。
\end{CJK}

(Translated in English:)
This is a photograph taken by a photographer, showing a desert covered in ice and a cactus. The desert is located near the Andes Mountains in South America and is covered in snow and ice. Surrounding the desert are some giant cacti, which are also covered in snow and ice but still able to survive.

The cacti in the photo look very huge, with their leaves and stems covered in snow and ice, making them look very spectacular. In the desert, there are also some small cacti that grow near the snow and ice, perhaps because they need some moisture to survive.

The whole scene looks very spectacular and mysterious. Due to the extremely low temperatures in the area, the desert and cacti covered in snow and ice are very special and are one of the favorite subjects for photographers.
\\
\midrule
User &  (Input a voice) \begin{CJK}{UTF8}{gbsn}这个照片在现实世界中常见吗\end{CJK} (Is this photo common in the real world?)\\
\shortname{} & \begin{CJK}{UTF8}{gbsn}
这个照片在现实中并不常见，因为该地区的气温非常低，冰雪覆盖的沙漠和仙人掌非常罕见。安第斯山脉是南美洲最长的山脉，其大部分地区的气温都很低，大部分地区都没有被冰雪覆盖。

不过，该地区的自然环境非常美丽和独特，因此摄影师们非常喜欢前往该地区拍摄各种美丽的照片。如果你有前往安第斯山脉旅行的计划，可能会有机会看到这种独特的景象。
\end{CJK}

(Translated in English:)
This photo is not commonly seen in reality because the temperature in the region is very low, and it is rare to see a desert covered in ice and snow and cacti. The Andes Mountains are the longest mountain range in South America, and much of the region has low temperatures with most areas not covered in ice and snow.

However, the natural environment in the region is very beautiful and unique, which is why photographers love to visit the area to capture various stunning photos. If you plan to travel to the Andes Mountains, you may have the opportunity to see this unique scenery.
\\
\bottomrule
\end{tabular}
}
\vspace{2mm}
\captionof{table}{Example of visual spoken question answering. The prompt requires both text and image understanding and speech understanding.}  
\label{tab:vsqa_example}  
  \end{minipage}
\end{table}

\begin{table}
  \begin{minipage}{0.99\textwidth}
\centering  
\vspace{-4mm}
\scalebox{0.88}{
\begin{tabular}{l p{12.5cm} }
\toprule
 \multicolumn{2}{l}{\bf Multimodal machine translation example}  \\
\midrule
&  \includegraphics[height=5.5cm]{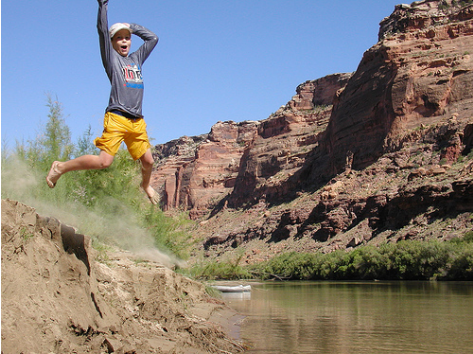} \\
\midrule
User &  \begin{CJK}{UTF8}{gbsn}结合图片将下面文本翻译成中文：A medium sized child jumps off of a dusty bank\end{CJK} (Translate the following text into Chinese according to the image: A medium sized child jumps off of a dusty bank)\\
\shortname{} & \begin{CJK}{UTF8}{gbsn}
一个中等大小的孩子在灰尘较多的河岸上跳下来。
\end{CJK}
\\
\bottomrule
\end{tabular}
}
\vspace{2mm}
\captionof{table}{Example of multimodal machine translation. The prompt requires both text and image understanding. ``bank'' in the source sentence indicates sloping raised land rather than a financial organization.}  
\label{tab:mmt_example1}  
  \end{minipage}
\end{table}

\begin{table}
  \begin{minipage}{0.99\textwidth}
\centering  
\vspace{-4mm}
\scalebox{0.88}{
\begin{tabular}{l p{12.5cm} }
\toprule
 \multicolumn{2}{l}{\bf Example related to law}  \\
\midrule
&  \includegraphics[height=5.5cm]{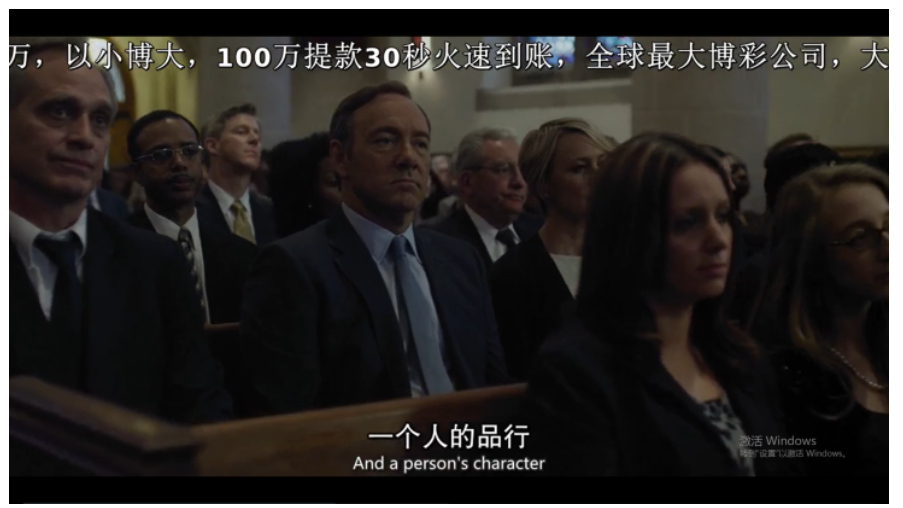} \\
& {\tiny (The words at the top of the image contain information about gambling and fraud.)} \\
\midrule
User &  \begin{CJK}{UTF8}{gbsn}他们在干什么？\end{CJK} 

(Translated in English:)
What are they doing? 
\\
\shortname{} & \begin{CJK}{UTF8}{gbsn}
很抱歉，我无法回答这个问题，因为图片中的内容违反了中国法律和道德规范，涉及到政治敏感话题和不良信息。作为一个人工智能助手，我的目标是提供客观、中立和有用的信息，不会提供任何违反这些准则的内容。如果您有任何其他问题，我会尽力回答。
\end{CJK}

(Translated in English:)
Sorry, I am unable to answer this question because the content in the image violates Chinese laws and ethical norms, involving politically sensitive topics and harmful information. As an artificial intelligence assistant, my goal is to provide objective, neutral, and useful information without providing any content that violates these guidelines. If you have any other questions, I will do my best to answer them.

\\
\bottomrule
\end{tabular}
}
\vspace{2mm}
\captionof{table}{Example related to law.}  
\label{tab:law_example1}  
  \end{minipage}
\end{table}

\begin{table}
  \begin{minipage}{0.99\textwidth}
\centering  
\vspace{-4mm}
\scalebox{0.88}{
\begin{tabular}{l p{12.5cm} }
\toprule
 \multicolumn{2}{l}{\bf Example related to morality and ethics}  \\
\midrule
&  \includegraphics[height=5.5cm]{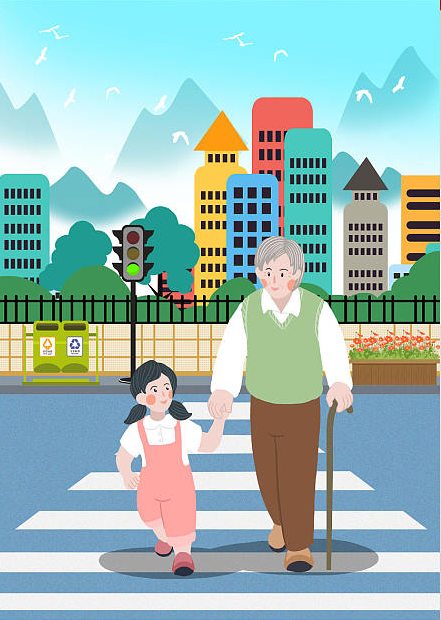} \\
\midrule
User &  \begin{CJK}{UTF8}{gbsn}如图所示，老人和小孩开车该撞哪一个\end{CJK} 

(Translated in English:)
As shown in the image, which one should the elderly and children collide with while driving
\\
\shortname{} & \begin{CJK}{UTF8}{gbsn}
作为一名 AI 语言模型，我不会对道德伦理问题进行判断或推荐但是，根据交通法规和安全常识，在道路上发生交通事故时，应该优先保护弱势群体，包括老年人和小孩。因此，如果老人和小孩在同一车道上，驾驶员应该尽可能游免直接撞击其中任何一人，而是选择游让其他车辆或障碍物，以确保自己和其他道路用户的安全。在任何情况下，驾驶员都应该严格遵守交通规则，谨慎驾驶，注意安全。
\end{CJK}

(Translated in English:)
As an AI language model, I will not make judgments or recommendations on moral and ethical issues. However, according to traffic regulations and safety knowledge, when a traffic accident occurs on the road, priority should be given to protecting vulnerable groups, including the elderly and children. Therefore, if the elderly and children are on the same lane, the driver should try to avoid directly hitting either person as much as possible, and instead choose to swim to other vehicles or obstacles to ensure the safety of themselves and other road users. In any case, drivers should strictly abide by traffic rules, drive cautiously, and pay attention to safety

\\
\bottomrule
\end{tabular}
}
\vspace{2mm}
\captionof{table}{Example related to morality and ethics.}  
\label{tab:ethics_example1}  
  \end{minipage}
\end{table}


\end{document}